\documentclass{article}

\usepackage{arxiv}

\usepackage[utf8]{inputenc}
\usepackage[T1]{fontenc}
\usepackage{hyperref}
\usepackage{url}
\usepackage{graphicx}
\usepackage{booktabs}
\usepackage{amsmath, amsfonts, amssymb}
\usepackage{natbib}
\usepackage{comment}
\newcommand{\Rey}{\mathrm{Re}}
\graphicspath{{./figures/}}

\title{Explainable deep reinforcement learning reveals energy-efficient control strategies for turbulent drag reduction}

\author{
Federica Tonti \\
Department of Aerospace Engineering\\
University of Michigan\\
Ann Arbor, MI 48109, United States\\
\texttt{ftonti@umich.edu}
\And
Ricardo Vinuesa \\
Department of Aerospace Engineering\\
University of Michigan\\
Ann Arbor, MI 48109, United States\\
\texttt{rvinuesa@umich.edu}
}

\begin{document}

\maketitle

\begin{abstract}
We propose a method combining Multi-Agent Deep Reinforcement
Learning (MARL) and eXplainable Deep Learning (XDL) to reduce drag in wall-bounded turbulent flows. Taking as a baseline the results of training agents directly targeting wall-shear stress and opposition control, three SHAP-guided approaches are compared. In the first, the
reward is computed from SHAP attributions of a U-net predicting the future velocity field; in the second, from SHAP attributions of a U-net predicting the skin-friction coefficient; in the third, from a combination of SHAP attributions of two U-nets predicting the skin-friction coefficient and the wall pressure fluctuations,
respectively. The combined SHAP strategy based on skin-friction coefficient and wall-pressure fluctuations achieves the best overall performance, achieving a DR of 34.44\% and a NES of 34.01\% with only 0.43\% normalized input power. Relative to opposition control, drag reduction and net energy saving increase by 49.41\% and 48.52\%, respectively. Compared with the direct wall-shear-stress baseline, the proposed strategy simultaneously improves performance while reducing the normalized actuation cost from 5.90\% to 0.43\%. Analysis of the
results reveals that the energetically efficient policy is consistent with pressure-gated actuation, activating predominantly at near-zero wall pressure, and operates on a temporal timescale comparable to the lifetime of the near-wall turbulent structures.
\end{abstract}


\section{Introduction}
\label{sec:intro}
Skin-friction drag in wall-bounded turbulence accounts for roughly a third of global transport energy consumption \citep{vinuesa2022brunton}, so even modest reductions yield significant benefits. The natural target of active flow control is the near-wall self-sustaining cycle, in which streamwise vortices generate low-speed streaks that become unstable and regenerate \citep{waleffe1997,jimenez1999,jimenez2018}. Classical strategies such as opposition control \citep{choi1994} disrupt this cycle but rely on an a-priori choice of the targeted structures and degrade with Reynolds number \citep{marusic2010}. A further constraint is the energetic cost of the actuation: as shown by \citet{bewley2009}, the actuation power can substantially reduce or even eliminate the power saved by drag reduction, so that canonical strategies achieving $20$--$40\%$ drag reduction in Direct Numerical Simulations (DNS) \citep{quadrio2004,quadrio2011,fukagata2024} often yield small or negative net energy saving \citep{gattiQuadrio2016}. \\
Deep reinforcement learning (DRL) has emerged as a model-free alternative for discovering nonlinear control policies in fluid systems \citep{suarez2025,vignon2023,font2025,tonti2025jcp,tonti2025combustion,guastoni2023,sonoda2023}, and multi-agent DRL (MARL) substantially outperforms opposition control in wall-bounded turbulence \citep{guastoni2023}; the resulting policies, however, are opaque about which structures drive the control. Explainable deep learning addresses this through attribution methods such as SHapley Additive exPlanations (SHAP), which has proven informative for turbulence \citep{lundberg2017,cremades2024,cremades2025coherent,cremades2025review,lozanoduran2021,osawa2024}. Crucially, \citet{hoyas2025skinfriction} showed that SHAP attributions for the skin-friction coefficient $c_f$ and for the velocity field identify qualitatively different structures, so a reward based on velocity-SHAP does not, in general, target the structures most responsible for drag. This is the limitation of the first XDL-guided DRL controller \citep{beneitez2025}, whose velocity-SHAP reward achieved the best drag reduction ($33.3\%$) among four strategies but identifies regions informative for the flow evolution rather than for skin friction. In this work we instead compute SHAP values from U-nets trained to predict the skin-friction coefficient $c_f$ and the wall-pressure fluctuation $p_w$, and use the resulting maps as the reward signals of a MARL controller, comparing five configurations: opposition control \citep{choi1994}, direct wall-shear stress reduction (WSE) \citep{guastoni2023}, velocity-SHAP (SHAP vel) \citep{beneitez2025}, SHAP $c_f$, and a combined SHAP $c_f+p_w$, all trained in a small channel configuration (SCC) and deployed in a large channel configuration (LCC) on 50 unseen initial conditions, with net energy saving \citep{kametani2011} as the main figure of merit.
Two promising approaches to drag reduction have so far been explored mostly independently: DRL policies that outperform classical control but at an actuation cost impacting the energy budget \citep{guastoni2023,beneitez2025}, and the energy-efficient pathway of \citet{marusic2021}, in which positive net energy saving is recovered by matching the actuation timescale to that of the targeted structures. We show that aligning the SHAP attribution target with the wall-friction dynamics, augmented with the wall-pressure fluctuation, reconciles the two: drag reduction (DR) outperforms the best DRL policies reported for this configuration, while the actuation cost drops considerably relative to direct wall-shear stress reduction, recovering the net energy saving (NES) of the DRL family at a level comparable to opposition control. This gain is consistent with pressure-gated actuation, in which the controller acts predominantly at near-zero wall pressure and on a timescale comparable to the lifetime of the near-wall structures. The gating is not built into the reward but emerges from the attribution target, identifying the attribution target as a previously unexploited design choice in XDRL-guided control: a transferable principle, aligning the target variable with the control objective, that is readily testable at higher Reynolds numbers and additional geometrical configurations. The numerical setup and DRL framework are described in \S\,\ref{sec:setup}, the SHAP-guided reward and configurations in \S\,\ref{sec:configurations_rew}, results in \S\,\ref{sec:results}, and conclusions in \S\,\ref{sec:conclusions}.

\section{Numerical setup and DRL framework}
\label{sec:setup}

\subsection{Flow configuration and DNS}
\label{sec:dns}
A turbulent open channel with friction Reynolds number $\Rey_\tau = u_\tau h / \nu = 180$ is considered, where $h$ is the channel height, $u_\tau = \sqrt{\tau_w / \rho}$ the friction velocity, and $\tau_w$ the mean wall-shear stress. We denote the streamwise, wall-normal and spanwise coordinates by $(x,y,z)$ and the corresponding velocity components by $\boldsymbol{u} = (u,v,w)$; viscous units are indicated by the superscript ${}^+$. The flow is scaled by $h$, velocities by the laminar centreline value $U_\mathrm{cl}$. The governing equations are
\begin{equation}
\partial_t \boldsymbol{u} + (\boldsymbol{u}\cdot\nabla)\boldsymbol{u} = -\nabla p + \Rey^{-1}\Delta \boldsymbol{u},\qquad \nabla\cdot\boldsymbol{u} = 0,
\label{eq:ns}
\end{equation}
where $p$ is a spatially uniform, time-dependent mean pressure gradient adjusted at each time step to maintain a constant bulk velocity $U_b = 2/3$. At the bottom wall, $u=w=0$ and the wall-normal velocity is prescribed as $v=v_\mathrm{bc}(x,z,t)$; the top boundary is a free-slip symmetry plane. Periodicity is enforced in $x$ and $z$. Training is carried out in a small channel configuration (SCC) of size $[L_x,L_y,L_z] = [2.67,\,1,\,0.8]$, large enough to support a single near-wall self-sustaining cycle \citep{jimenezmoin1991,waleffe1997,jimenez1999}. Policy inference is performed in a large channel configuration (LCC) of size $[L_x,L_y,L_z] = [2\pi,\,1,\,\pi]$. The open-source spectral solver \texttt{Dedalus} \citep{burns2020dedalus} is used, with Fourier expansions along $x$ and $z$ and Chebyshev polynomials along $y$. The SCC resolution is $[N_x,N_y,N_z]=[16,\,64,\,16]$ grid points and the LCC $[64,\,64,\,32]$, in line with the convergence assessments reported for this Reynolds number by \citet{guastoni2023} and \citet{beneitez2025}. Time integration uses the third-order four-step Runge-Kutta scheme of \citet{ascher1997} with a fixed step $\Delta t^+ \approx 0.039$. After an initial transient, the flow reaches a statistically stationary state. A database of uncontrolled snapshots covering $t^+ > 2 \times 10^4$ is collected in the SCC to train a U-net that, given the velocity field at time $t$, predicts the target variable at $t+\Delta t$, with $\Delta t^+ = 5$. SHAP values are then computed, and a second surrogate U-net is trained to map the velocity field at time $t$ to the SHAP field at the same instant. Further details are given in \citet{beneitez2025}.

\subsection{Multi-agent deep reinforcement learning}
\label{sec:drl}
Control is applied to the bottom wall by blowing and suction, with the wall-normal velocity $v_\mathrm{bc}(x,z,t)$ as the actuation field. To preserve the bulk flow rate, the spatial mean of the prescribed actuation is subtracted at every step so that the net mass flux through the wall remains identically zero. The actuation amplitude is limited by $|v_\mathrm{bc}| \leq u_\tau$, a bound consistent with previous DRL studies on the same flow \citep{guastoni2023,sonoda2023,beneitez2025}. Rather than assigning each agent a finite-area patch of the wall, the  structure of the spectral discretization is exploited: one agent is placed at every collocation point in the homogeneous directions, giving $N_a=16\times16=256$ agents during SCC training and run time evaluation. At inference, the shared policy is applied pointwise to $64\times32$ wall locations of the LCC. All agents share a single policy network, so each control step provides $N_a$ independent trajectories along which the gradient is computed. Every agent receives as local observation the pair $(u',v')$ of streamwise and wall-normal velocity fluctuations sampled at the sensing plane $y^+=15$, with the instantaneous spatial average removed to avoid spurious drifts between controlled and uncontrolled rollouts. The twin-delayed deep deterministic policy-gradient (TD3) algorithm \citep{fujimoto2018td3} is adopted through Stable-Baselines3 \citep{raffin2021stablebaselines3}, and the DRL loop is coupled to the DNS code through a Gymnasium environment that wraps the solver and exposes states, actions and rewards. Actions are taken every $\Delta t^+_\mathrm{ctrl}\approx 0.5$, much longer than the DNS step but still short compared to the WSE-turnover time at $y^+=15$; the instantaneous reward is accumulated over the DNS substeps and the agents receive its time average. An episode comprises $3 \times 10^3$ interactions and starts from one of six instantaneous snapshots randomly drawn from the uncontrolled database, with exploration encouraged through Gaussian action noise of amplitude $0.1\,u_\tau$. During training, the policy is periodically evaluated on a separate environment that uses the same solver and reward computation but independent initial conditions, and the model with the highest mean reward is retained. The statistics reported in the results are computed on 50 LCC rollouts from a further unseen set of initial conditions, providing an out-of-sample test of each policy.

\subsection{SHAP-guided configurations and reward formulation}
\label{sec:configurations_rew}
The SHAP-guided controllers are obtained by varying the predictive target $Y$ of the auxiliary U-net from which SHAP attributions are extracted. The input is the velocity flow field at time $t$, $\boldsymbol{u}_t$; the targets $Y$ are the future velocity field, $Y = \boldsymbol{u}_{t+\Delta t}$, the skin-friction coefficient, $Y = c_{f_{{t+\Delta t}}}$, and the wall pressure fluctuations, $Y = p_{w_{{t+\Delta t}}}$. Three configurations are considered: SHAP vel, in which $Y$ is the future velocity field, reproduces the work of \citet{beneitez2025}; SHAP $c_f$ replaces the predictive target with $c_f$; SHAP $c_f + p_w$ uses a weighted combination of $c_f$ and $p_w$. In every case the DRL agent and the predicting U-net are trained in the SCC and subsequently deployed in the LCC at inference, on the same set of 50 unseen initial conditions, so that the resulting DR, actuation power and NES \citep{hasegawa2014} can be fairly compared. The MARL framework described in \S\,\ref{sec:drl} is shared across all simulations, and only the reward signal returned to the agents at every control step is modified. Two reward families are used. The baseline policy, denoted WSE and following \citet{guastoni2023,beneitez2025}, is trained directly on the wall-shear stress: at every control step the instantaneous skin friction is averaged over the actuator footprint and fed back to the corresponding agent with a negative sign, so that minimizing drag coincides with maximizing the reward. The three SHAP-guided policies share, instead, a single attribution-based reward,
\begin{equation}
  r(t) \;=\; -\,\frac{1}{|\Omega|}
  \int_{\Omega}\,
  \bigl\|\boldsymbol{\phi}(\boldsymbol{x},t)\bigr\|\,\mathrm{d}\Omega,
  \qquad
  \bigl\|\boldsymbol{\phi}\bigr\| \;=\; \sqrt{\phi_x^{2} + \phi_y^{2} + \phi_z^{2}},
  \label{eq:reward-shap}
\end{equation}
where $\Omega$ is the computational domain and $\boldsymbol{\phi}(\boldsymbol{x},t)$ is the SHAP attribution vector field, which quantifies the pointwise contribution of the local flow state to the surrogate-model prediction of the target quantity (here $c_f$ or $p_w$). Its magnitude $\|\boldsymbol{\phi}\|$ measures the influence of each spatial location irrespective of sign: a positive attribution increases the prediction and a negative one lowers it, but both indicate an active coherent structure that the surrogate deems relevant. Minimizing the domain-averaged magnitude therefore drives the policy toward flow states in which no region strongly contributes to the predicted quantity, effectively suppressing the near-wall vortical structures that sustain the near-wall cycle. In practice, evaluating $\boldsymbol{\phi}$ through gradient-SHAP at every control step would be extremely expensive, so the attribution is produced by a surrogate U-net that maps instantaneous velocity fluctuations directly to $\boldsymbol{\phi}(\boldsymbol{x},t)$ in a single forward pass within the DRL loop \citep{beneitez2025}. The reward is normalized by the domain area $|\Omega|$ and, during training, scaled by a reference value $r_{\mathrm{ref}}$ estimated from uncontrolled runs so that the signal is $O(1)$. Between two consecutive agent interactions, $r(t)$ is evaluated at every DNS substep and time-averaged before being passed to the controllers, so that the reward reflects the cumulative effect of each action over the corresponding control window; the same convention is applied to the WSE reward. The three SHAP-guided configurations use identical reward functionals \eqref{eq:reward-shap} and differ only in the predictive target $Y$ of the auxiliary U-net. For SHAP $c_f + p_w$, the attribution surrogate $g_\phi$ entering \eqref{eq:reward-shap} is obtained by merging two independently trained surrogate networks in parameter space,
\begin{equation}
  \theta^{\mathcal{C}^+}
  \;=\; \lambda\,\theta^{c_f}
  \;+\; (1-\lambda)\,\theta^{p_w},
  \qquad \lambda \in [0,1],
  \label{eq:param-merge}
\end{equation}
where $\theta^{c_f}$ and $\theta^{p_w}$ are the weight vectors of the surrogates trained on the $c_f$ and $p_w$ attribution fields, respectively. This operation is well-defined because both surrogates share the same architecture, and parameter-space interpolation between networks trained on different tasks has been shown to produce valid, well-posed model combinations \citep{wortsman2022}. The merged network $g_\phi^{\mathcal{C}^+}$ is queried at every control step to produce a single attribution field $\boldsymbol{\phi}^{\mathcal{C}^+}$, which enters \eqref{eq:reward-shap} directly. The weight $\lambda$ controls the relative contribution of the wall-shear-stress and wall-pressure attribution surrogates; we explored $\lambda \in [0.70, 0.95]$, with DR ranging from 33.5\% to 34.4\% and NES from 32.1\% to 34.0\%. The configuration reported throughout \S\,\ref{sec:results} corresponds to $\lambda = 0.8$, which achieved the highest NES and drag reduction.

\section{Results}
\label{sec:results}
The five policies are evaluated in the LCC on the same set of 50 initial conditions from an uncontrolled flow field, and all statistics are computed in the stationary regime, $t^+ > 500$. The percentages of DR and NES refer to uncontrolled flow.

\subsection{Performance}
\label{sec:results:performance}
\begin{figure}
  \centering
  \includegraphics[width=0.49\textwidth]{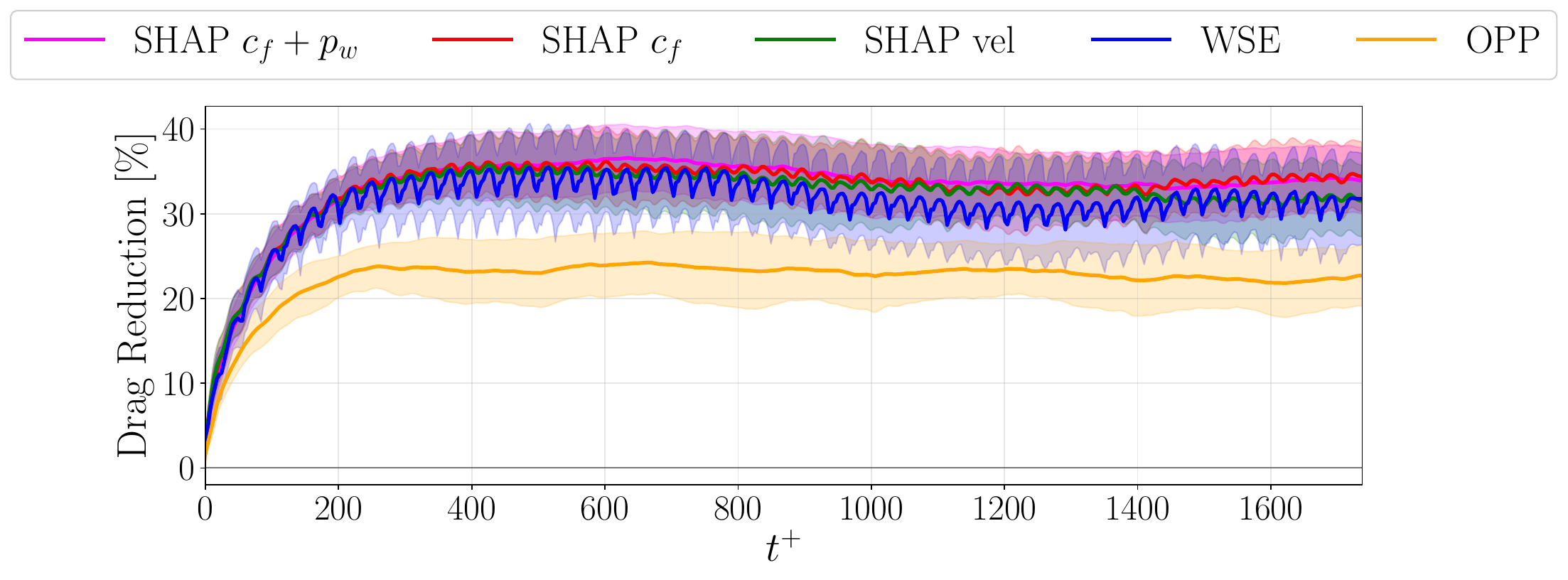}
  \hfill
  \includegraphics[width=0.49\textwidth]{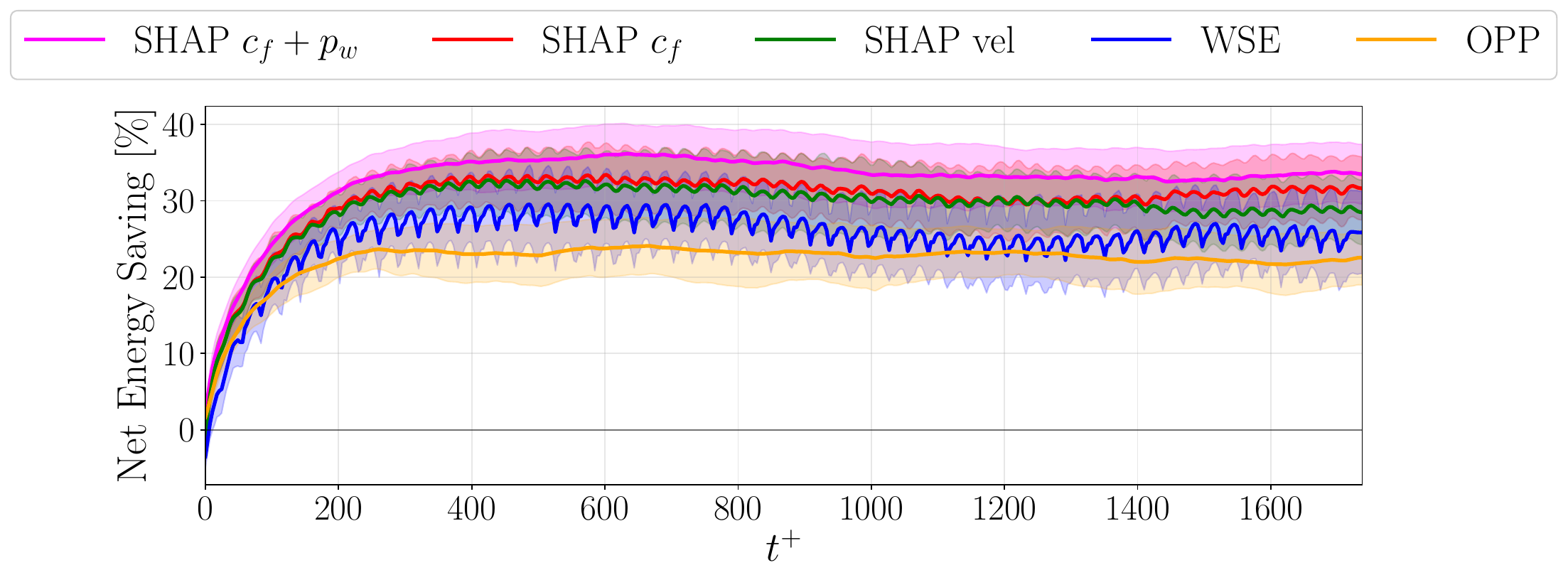}
  \caption{Drag reduction (left) and net energy
    saving (right) as a function of $t^+$ for the five policies with respect to the uncontrolled flow: SHAP vel (green), where the target is the velocity field, SHAP $c_f$ (red) where the target is $c_f$, SHAP $c_f + p_w$ (magenta) which considers a weighted combination of $c_f$ and $p_w$, WSE (blue) where the algorithm is trained directly on the wall-shear stress and OPP (orange) is the opposition control. Solid lines are the mean across the 50 LCC
    rollouts and shaded envelopes show the standard deviation.}
  \label{fig:ensemble}
\end{figure}

Figure~\ref{fig:ensemble} shows the DR and NES over the rollouts, and Table~\ref{tab:performance} reports the stationary-regime averages of each configuration in absolute terms and relative to the opposition-control (OPP) baseline of \citet{choi1994}, together with the decomposition of the actuation cost
\begin{equation}
w_\mathrm{in} = \langle p_w\, v_w\, S_1 \rangle+\left\langle \tfrac{1}{2} v_w^3\, S_2 \right\rangle,
\label{eq:win}
\end{equation}
into its pressure-work and kinetic components, where $S_1 = 1$ if $p_w v_w > 0$ and zero otherwise, and $S_2 = 1$ if $v_w > 0$ and zero otherwise \citep{hasegawa2014}. Only the positive part is retained, so that $w_\mathrm{in}$ measures the net energy input to the flow. The skin-friction coefficient $c_f = 2\nu\langle\partial u/\partial y\rangle_{x,z,\mathrm{wall}} / U_b^2$ is plane-averaged, and wall-normal velocity and wall pressure are reported in viscous units, $v_w^+ = v_w / u_\tau$ and $p_w^+ = p_w / u_\tau^2$, with $u_\tau = \sqrt{\tau_{w_0}/\rho}$ the friction velocity of the uncontrolled flow. The DR and NES are defined as
\begin{equation}
\mathrm{DR} = \frac{c_{f_0} - \langle c_f \rangle}{c_{f_0}}, \qquad
\mathrm{NES} = \mathrm{DR} - \frac{w_{\mathrm{in}}}{c_{f_0}},
\label{eq:dr-nes}
\end{equation}
where $c_{f_0} = 2\tau_{w_0} / U_b^2$ is the skin-friction coefficient of uncontrolled flow and $\langle c_f \rangle$ the skin-friction coefficient averaged over the wall plane and over time at stationary regime.

\noindent Three observations follow. First, all four reinforcement-learning policies achieve higher DR than opposition control ($23.0\%$): WSE reaches $31.9\%$, SHAP vel $33.2\%$, SHAP $c_f$ $34.1\%$, and SHAP $c_f + p_w$ $34.4\%$. Switching the predictive target of the auxiliary U-net from the velocity field to $c_f$ thus improves DR by roughly one percentage point, with a further small gain when the wall-pressure fluctuation is added. Second, the actuation cost varies by more than an order of magnitude across the four DRL policies. The viscous component $w_\mathrm{in}^{v}/c_{f_0}$ remains below $0.15\%$ in every configuration, so the cost is dominated by the pressure-work term: WSE shows $5.7\%$ of $c_{f_0}$, SHAP vel and SHAP $c_f$ reduce this to $2.9\%$ and $2.73\%$, and SHAP $c_f + p_w$ reaches $0.4\%$, roughly seven times smaller than SHAP vel and fourteen times smaller than WSE. Opposition control attains the lowest cost of any policy ($0.1\%$) but the lowest DR. Third, this reduction translates directly into NES, which increases from $22.9\%$ for opposition control to $34.0\%$ for SHAP $c_f + p_w$: the combined model recovers the NES of opposition control while retaining the higher DR of the DRL family.

\begin{table}
  \centering
  \setlength{\tabcolsep}{4pt}
  \resizebox{\textwidth}{!}{%
  \begin{tabular}{lccccccccc}
    Case & DR [\%] & $\Delta$DR [pp] & rel-DR [\%] & $w_\mathrm{in}^{p}/c_{f_0}$ [\%] & $w_\mathrm{in}^{v}/c_{f_0}$ [\%] & $w_\mathrm{in}/c_{f_0}$ [\%] & $NES$ [\%] & $\Delta NES$ [pp] & rel-$NES$ [\%] \\
    \hline
    SHAP $c_f + p_w$ & $34.44\pm 1.74$ & $+11.39$ & $49.41$ & $0.41 \pm 0.05$ & $0.02 \pm 0.00$ & $0.43 \pm 0.05$ & $34.01 \pm 1.74$ & $+11.11$ & $48.52$ \\
    SHAP $c_f$       & $34.12\pm 2.13$ & $+11.07$ & $48.03$ & $2.73 \pm 0.10$ & $0.03 \pm 0.00$ & $2.76 \pm 0.10$ & $31.36 \pm 2.20$ & $+8.46$  & $36.94$ \\
    SHAP vel         & $33.20\pm 2.06$ & $+10.15$ & $44.04$ & $2.98 \pm 0.05$ & $0.04 \pm 0.00$ & $3.02 \pm 0.05$ & $30.18 \pm 2.07$ & $+7.28$  & $31.79$ \\
    WSE              & $31.94\pm 2.56$ & $+8.89$  & $38.57$ & $5.76 \pm 0.35$ & $0.14 \pm 0.00$ & $5.90 \pm 0.35$ & $26.05 \pm 2.24$ & $+3.15$  & $13.76$ \\
    OPP              & $23.05\pm 1.58$ & $0.00$   & $0.00$  & $0.13 \pm 0.01$ & $0.02 \pm 0.00$ & $0.15 \pm 0.02$ & $22.90 \pm 1.59$ & $0.00$   & $0.00$  \\
  \end{tabular}}
  \caption{Performance of the five control methods, evaluated on $50$ LCC rollouts for $t^+ > 500$. DR, pressure and viscous components of the actuation cost ($w_\mathrm{in}^{p}, w_\mathrm{in}^{v}$), total actuation cost ($w_\mathrm{in}$), and NES are reported with respect to the uncontrolled flow; DR and NES are normalized by $c_{f_0}$ so that they can be expressed as percentages. $\Delta$DR and $\Delta NES$ are the absolute differences in percentage points relative to the opposition-control (OPP) baseline, and rel-DR and rel-$NES$ are the corresponding relative improvements over the OPP reference. All the models except OPP are trained in a SCC and deployed in the LCC. Uncertainties denote variability over initial conditions.}
  \label{tab:performance}
\end{table}

\subsection{Near-wall structures and actuation}
\label{sec:results:structures}
\begin{figure}
  \centering
  \includegraphics[width=\textwidth]{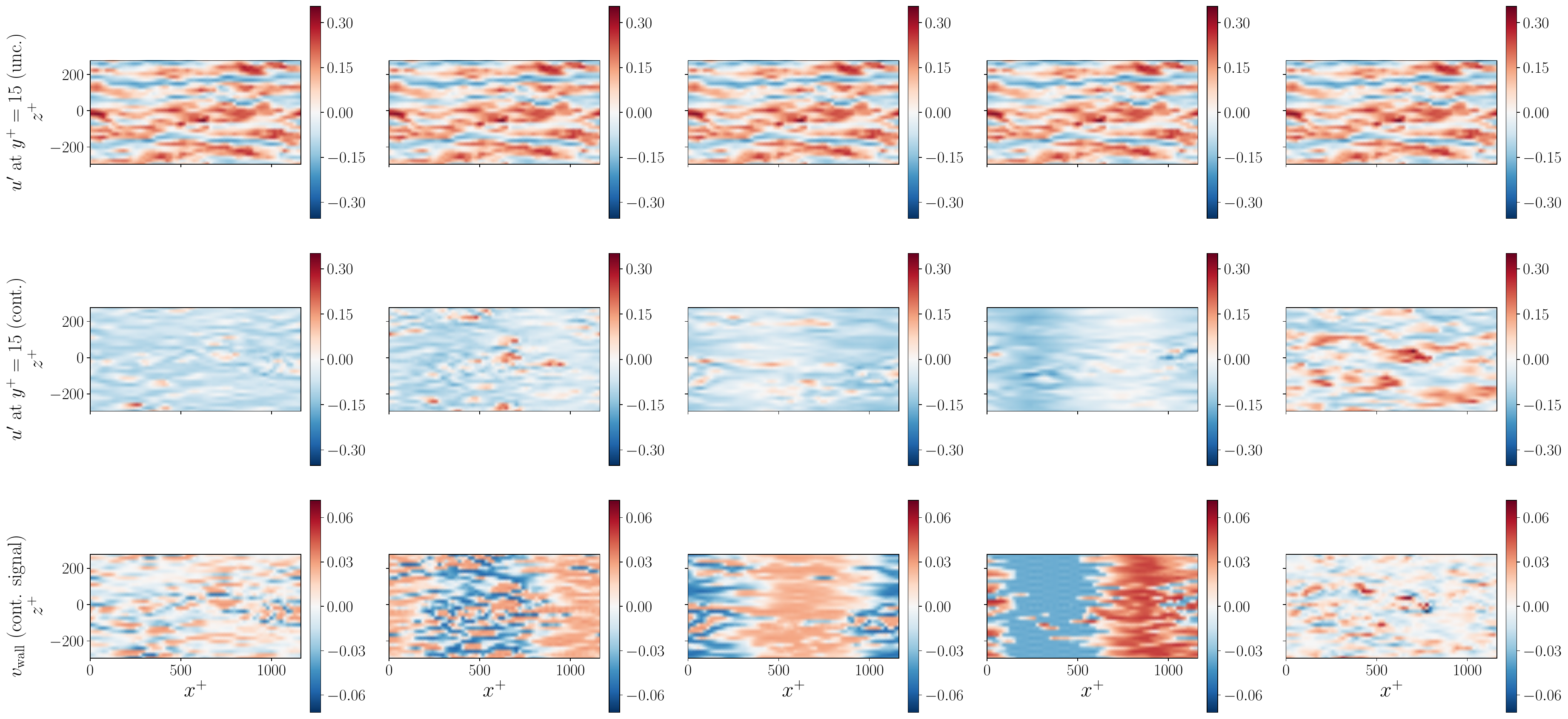}
  \caption{Instantaneous near-wall fields for the five control methods. Top row: streamwise velocity fluctuation $u'$ at $y^+ = 15$ in the uncontrolled flow. Middle row: $u'$ at $y^+ = 15$ under control. Bottom row: wall-normal control signal $v_\mathrm{wall}$. All snapshots are selected from the same instant of a representative rollout. Columns from left to right: SHAP $c_f + p_w$, SHAP $c_f$, SHAP vel, WSE, OPP. The structural reorganization of $u'$ at the sensing plane is comparable across the four reinforcement-learning policies, while the wall-normal control signal differs: WSE, SHAP vel and SHAP $c_f$ actuate in large streamwise-coherent patches, whereas SHAP $c_f + p_w$ and OPP produce small-scale low-amplitude actuation throughout the domain.}
  \label{fig:fields}
\end{figure}

Figure~\ref{fig:fields} compares the instantaneous streamwise velocity fluctuation $u'$ at $y^+ = 15$ in the uncontrolled flow (top row) and under each of the five policies (middle row), together with the corresponding wall-normal control signal $v_\mathrm{wall}$ (bottom row). The four DRL policies reduce the streaks, lowering the amplitude of $u'$ at $y^+ = 15$ and reorganizing the streak pattern, whereas opposition control suppresses them only partially, leaving residual streamwise-coherent structures across the domain, consistent with its smaller drag reduction. The quadrant decomposition of the Reynolds-stress events $(u', v')$ at the sensing plane is reported in Table~\ref{tab:quadrant}. In the uncontrolled flow, sweeps (Q4) occupy $54.0\%$ and ejections (Q2) $30.2\%$ of the joint probability, consistent with the standard near-wall energetics \citep{jimenezmoin1991,hoyas2025skinfriction}. All four DRL policies suppress Q4 to between $2.3\%$ (WSE) and $11.3\%$ (SHAP $c_f + p_w$) and amplify Q2 to around $50\%$, while opposition control preserves a distribution close to the uncontrolled flow ($48.8\%$ in Q4, $39.1\%$ in Q2), in line with its action being concentrated on the wall-normal velocity rather than on the streak-sweep cycle. The wall-normal control signal $v_\mathrm{wall}$ in Figure~\ref{fig:fields} reveals where the policies differ. WSE and SHAP vel actuate in large streamwise-coherent patches of one sign that extend over hundreds of viscous units in $x^+$, with $v_\mathrm{wall}$ saturating close to the amplitude limit; SHAP $c_f$ produces patches of intermediate extent at comparable amplitude; and SHAP $c_f + p_w$ and opposition control generate small-scale, low-amplitude actuation throughout the domain, with no streamwise organization visible. The same structural reorganization at the sensing plane is achieved through very different actuation cost, with the distinguishing signal contained in the actuation field itself rather than in the second-order flow statistics.

\begin{table}
  \centering
  \begin{tabular}{lcccc}
    Case & Q1 [\%] & Q2 (eject) [\%] & Q3 [\%] & Q4 (sweep) [\%] \\
    \hline
    Uncontrolled          & 12.2 & 30.2 & 3.6  & 54.0 \\
    SHAP $c_f + p_w$ & 1.2  & 53.0 & 34.4 & 11.3 \\
    SHAP $c_f$            & 0.7  & 50.6 & 43.8 & 4.9  \\
    SHAP vel              & 0.9  & 50.4 & 42.8 & 5.9  \\
    WSE                   & 0.5  & 51.5 & 45.6 & 2.3  \\
    OPP                   & 6.8  & 39.1 & 5.3  & 48.8 \\
  \end{tabular}
  \caption{Occupancy fractions of the joint probability of
    $(u', v')$ at $y^+ = 15$ for the uncontrolled flow and the
    five controlled cases.}
  \label{tab:quadrant}
\end{table}

\subsection{Mechanism: pressure-gated actuation}
\label{sec:results:mechanism}
\begin{figure}
  \centering
  \includegraphics[width=0.8\textwidth]{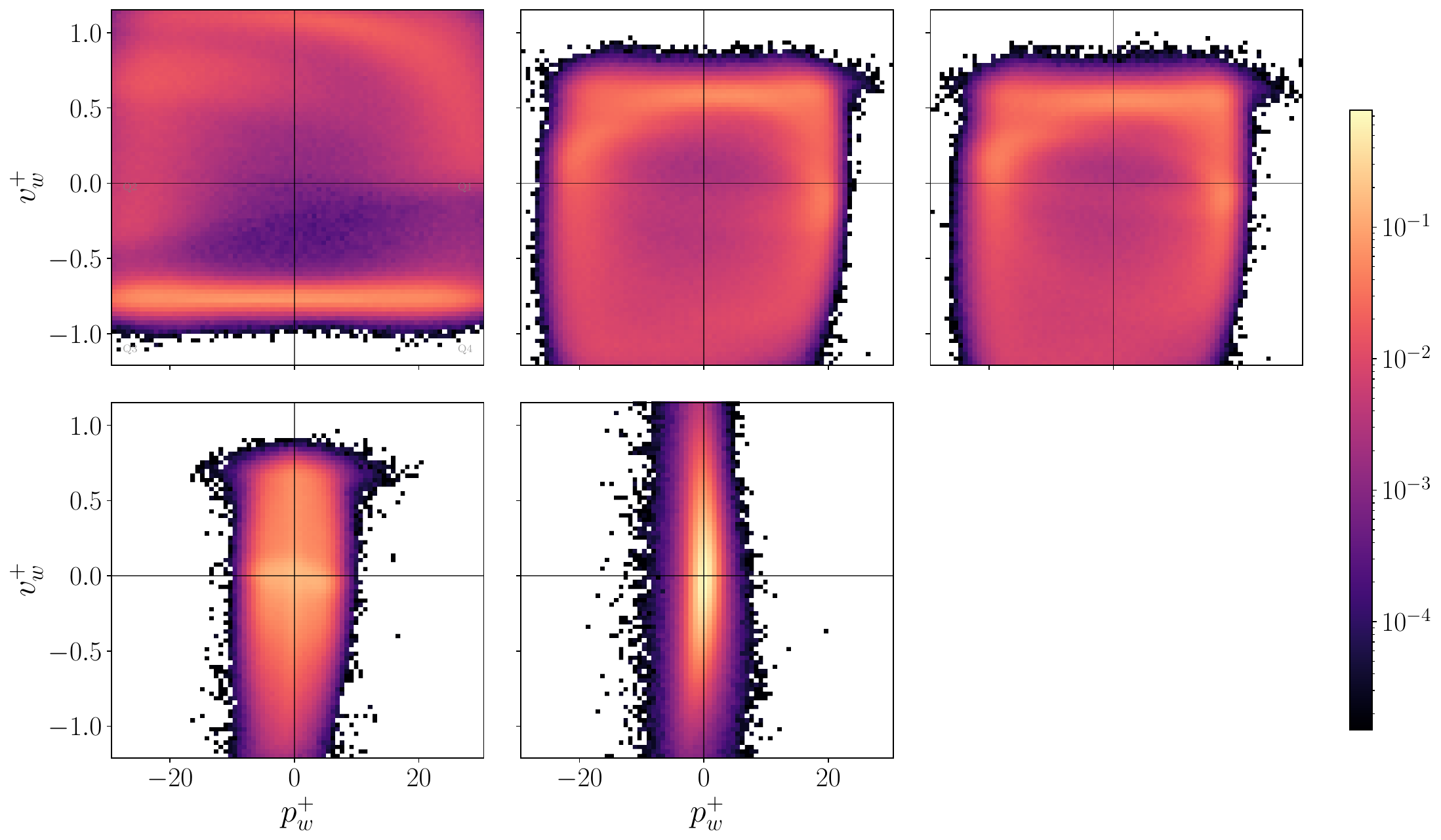}
  \caption{Joint probability density of the wall-pressure
    fluctuation $p_w^+$ and the wall-normal actuation $v_w^+$,
    computed pointwise over the wall and in the stationary regime
    ($t^+ > 500$) for the five policies. Each panel reports the time-averaged positive part of the pressure-work cost,
    $\langle [p_w v_w]^+ \rangle / c_{f_0}$, as a percentage of the uncontrolled friction coefficient. WSE, SHAP vel and SHAP $c_f$ distribute over the full range of $p_w^+$ with the actuation saturated near $v_w^+ \approx \pm 1$. SHAP $c_f + p_w$ and OPP focus on a narrow band at
    $p_w^+ \approx 0$, with the actuation amplitude reduced.}
  \label{fig:pwvw}
\end{figure}

\begin{figure}
  \centering
    \includegraphics[width=0.49\textwidth]{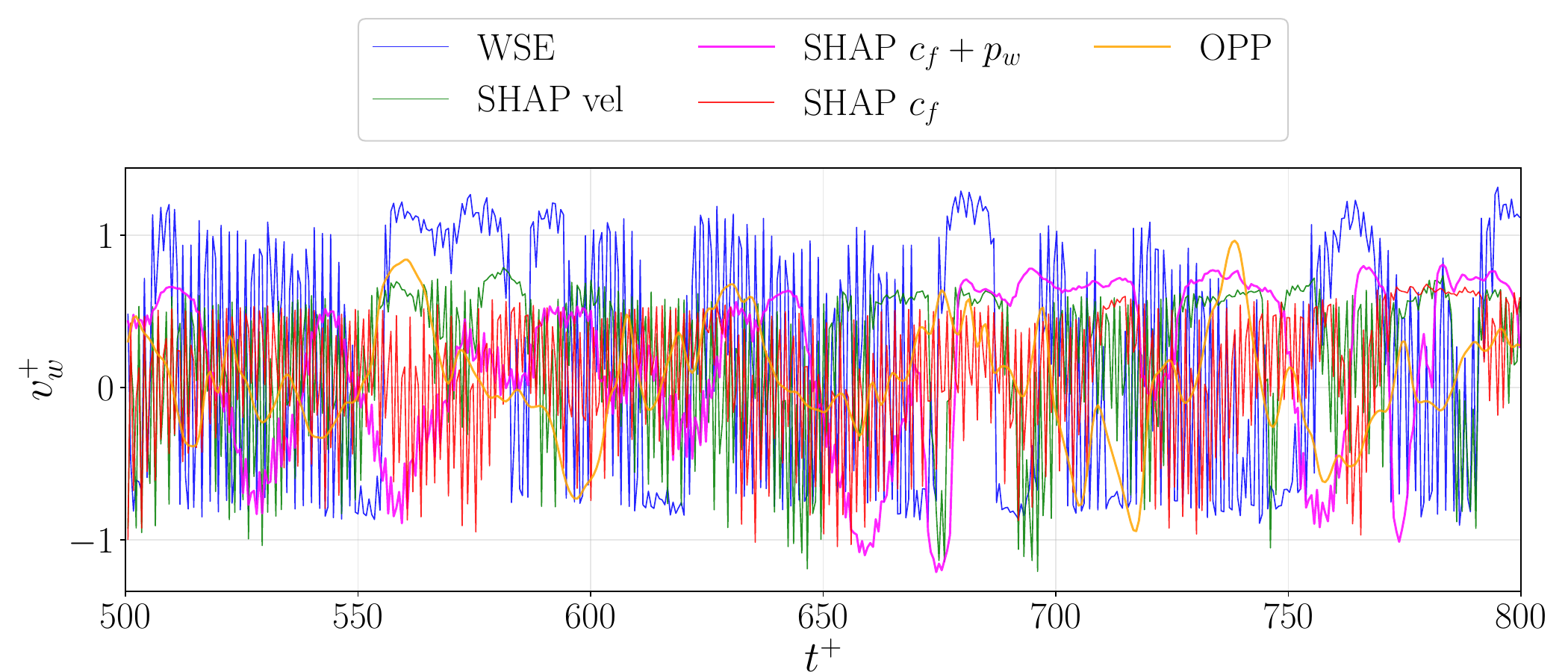}
    \hfill
    \includegraphics[width=0.49\textwidth]{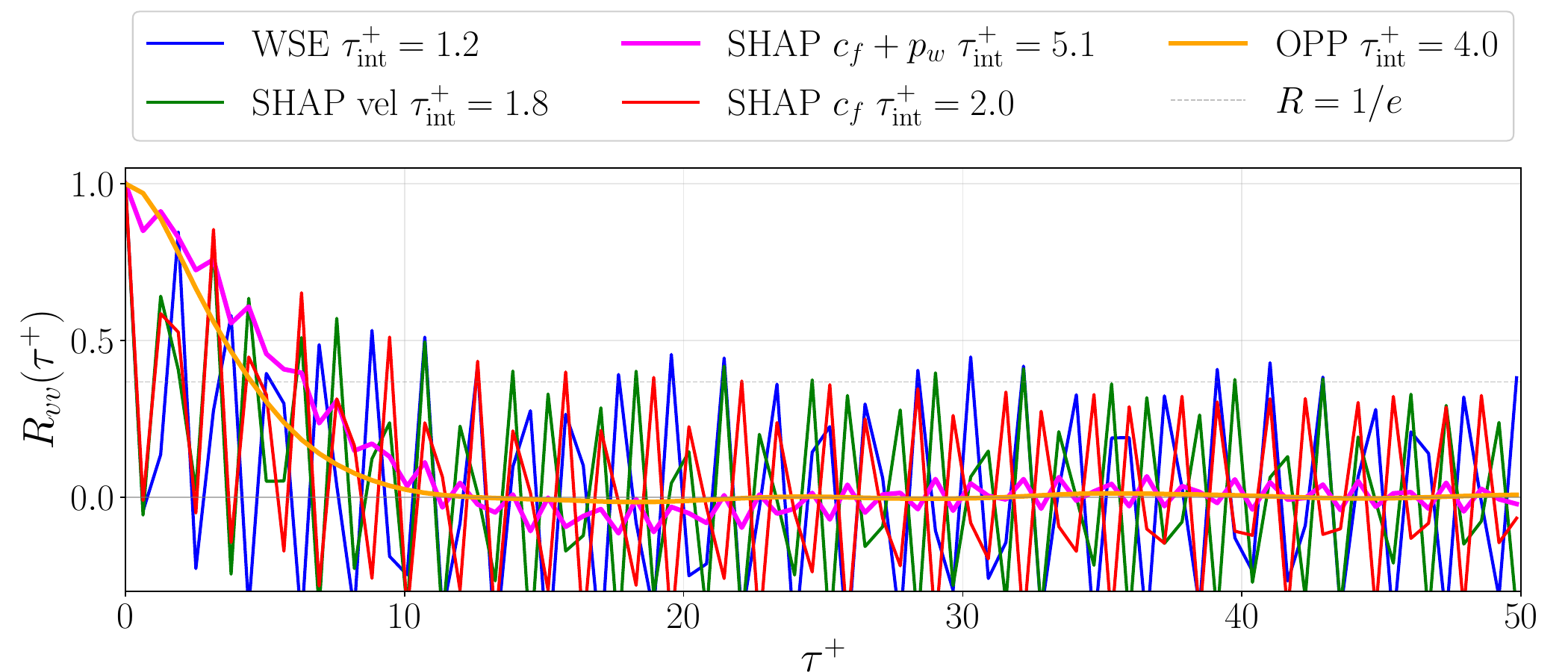}
    \caption{Temporal character of the wall-normal actuation in the stationary regime ($t^+ > 500$). Left: representative time series of $v_w^+$ at a single actuator location for the five policies. Right:
  temporal autocorrelation $R_{vv}(\tau^+)$ of $v_w$, averaged over the 50 LCC rollouts. The legend reports the integral time scale $\tau_{\mathrm{int}}^+ = \int_0^{\tau_0} R_{vv}\,\mathrm{d}\tau$.}
    \label{fig:temporal}
  \end{figure}

Since the actuation cost is dominated by the wall pressure-work term $\langle [p_w v_w]^+ \rangle$ in every configuration (Table~\ref{tab:performance}), Figure~\ref{fig:pwvw} shows the joint probability distribution of $p_w^+$ and $v_w^+$ over the wall at stationary regime. WSE, SHAP vel, and SHAP $c_f$ distribute across the full range $p_w^+ \in [-30, 30]$, with the actuation concentrated at the saturation amplitudes $v_w^+ \approx \pm 1$ and a pressure-work cost $\langle [p_w v_w]^+ \rangle / c_{f_0}$ between $2.7$ and $5.8\%$. SHAP $c_f + p_w$ and opposition control instead shows a narrow band centered at $p_w^+ \approx 0$, only up to $|p_w^+|\lesssim 10$. The actuation amplitude is reduced to $v_w^+ \approx 0$, dropping the cost to $0.41\%$ and $0.13\%$, respectively. Wall-pressure fluctuations themselves retain a comparable range across all policies; what is broken is the correlation between pressure and actuation. We interpret this as \emph{pressure-gated actuation}: the policy actuates predominantly when the local wall pressure is small. For SHAP $c_f + p_w$ this gating is not imposed in the reward, Eq.~\eqref{eq:reward-shap} being identical to the other SHAP configurations except for the inclusion of SHAP for $p_w$. This is an emergent property of the attribution target rather than a tailored regularizer. Figure~\ref{fig:temporal} shows the temporal character of the actuation: a representative time series of $v_w^+$ of a single actuator (left) and the autocorrelation $R_{vv}(\tau^+)$ averaged over 50 LCC rollouts (right). The integral time scale $\tau_{\mathrm{int}}^+ = \int_0^{\tau_0} R_{vv}(\tau)\,\mathrm{d}\tau$ is reported in the legend, with $\tau_0$ taken as the first lag at which $R_{vv}$ remains non-positive for three consecutive samples, which filters the spurious zero crossings of the noisier policies and yields a robust estimate. WSE, SHAP vel, and SHAP $c_f$ produce rapidly oscillating signals with $\tau_{\mathrm{int}}^+ \approx 1.2$--$2.0$, only marginally longer than the control update interval. SHAP $c_f + p_w$ and opposition control yield smooth signals with $\tau_{\mathrm{int}}^+ = 5.1$ and $4.0$, roughly three times larger, approaching the characteristic lifetime of near-wall quasi-streamwise vortices \citep[$\tau^+\sim 10$--$30$;][]{jimenez1999}. This similarity is physically significant: opposition control attains low $w_{\mathrm{in}}/c_{f_0}$ by construction, mirroring the sensed velocity and inheriting the temporal coherence of the structures it opposes. SHAP $c_f + p_w$ converging to a comparable time scale without any such built-in mechanism indicates that its pressure-aware model has guided the agent toward the same strategy, acting on the time scale of the structures responsible for drag rather than at every control step. The energetically-efficient policies therefore share two signatures, low wall-pressure work (Figure~\ref{fig:pwvw}) and slow, structure-locked actuation, both signatures of a control that targets the regeneration cycle rather than its instantaneous footprint.

\section{Conclusions}
\label{sec:conclusions}

We introduced a SHAP-guided multi-agent DRL framework for turbulent drag reduction in which the auxiliary U-net used to generate the attribution map is trained to predict wall quantities directly relevant to the control objective, rather than the future velocity field. Two configurations are tested at $\Rey_\tau = 180$ in an open-channel flow: SHAP $c_f$, with attributions from a U-net that predicts the skin-friction coefficient, and SHAP $c_f + p_w$, which combines attributions from U-nets predicting $c_f$ and $p_w$. The predictive U-net and the DRL agent are both trained in the small channel and evaluated in the large channel on 50 unseen initial conditions. Aligning the SHAP attribution with the control objective improves both drag reduction and net energy saving over SHAP vel, with the combined SHAP $c_f + p_w$ configuration giving the largest gains. The combined SHAP policy reduces actuation cost by a factor of about $14$ relative to WSE and about $7$ relative to SHAP vel, while OPP remains lower cost but achieves substantially lower drag reduction. SHAP $c_f + p_w$ recovers the energetic efficiency of opposition control while retaining the higher drag reduction of the DRL family. The reduction in actuation cost reflects a qualitatively distinct operating regime: the controller actuates predominantly at near-zero wall pressure and on a temporal timescale comparable to that of the underlying near-wall structures. \\
Future work will address whether the training domain, in addition to the attribution target, influences the performance of the XDRL strategy, and whether the same design principle transfers to higher Reynolds numbers.

\small\section*{Acknowledgments}
The authors gratefully acknowledge M. Beneitez for
the SHAP-guided multi-agent DRL framework, which served as the starting point for the present work. We also thank A. Cremades for helpful discussions on the SHAP-attribution methodology.

\section*{Declaration of interests}
The authors report no conflict of interest.

\bibliographystyle{unsrtnat}
\bibliography{references}

\end{document}